\documentclass[aps,prx,twocolumn,superscriptaddress]{revtex4-2}
\usepackage{amssymb}
\usepackage{amsfonts}
\usepackage{color}
\usepackage{graphicx}
\usepackage{amsmath}
\usepackage{times}
\usepackage{bm}
\usepackage{float}
\usepackage{lipsum}
\usepackage{import}
\usepackage{enumerate}
\begin{document}
\title{Inferring Attracting Basins of Power System with Machine Learning}
\author{Yao Du}
\thanks{These authors contributed equally to this work}
\affiliation{School of Physics and Information Technology, Shaanxi Normal University, Xi'an 710062, China}
\author{Qing Li}
\thanks{These authors contributed equally to this work}
\affiliation{School of Physics and Information Technology, Shaanxi Normal University, Xi'an 710062, China}
\author{Huawei Fan}
\affiliation{School of Science, Xi'an University of Posts and Telecommunications, Xi'an 710121, China}
\author{Meng Zhan}
\affiliation{School of Electrical and Electronic Engineering, Huazhong University of Science and Technology, Wuhan 430074, China}
\author{Jinghua Xiao}
\affiliation{School of Science, Beijing University of Posts and Telecommunications, Beijing 100876, China}
\author{Xingang Wang}
\email{E-mail address: wangxg@snnu.edu.cn}
\affiliation{School of Physics and Information Technology, Shaanxi Normal University, Xi'an 710062, China}

\begin{abstract}
Power systems dominated by renewable energy encounter frequently large, random disturbances, and a critical challenge faced in power-system management is how to anticipate accurately whether the perturbed systems will return to the functional state after the transient or collapse. Whereas model-based studies show that the key to addressing the challenge lies in the attracting basins of the functional and dysfunctional states in the phase space, the finding of the attracting basins for realistic power systems remains a challenge, as accurate models describing the system dynamics are generally unavailable. Here we propose a new machine learning technique, namely balanced reservoir computing, to infer the attracting basins of a typical power system based on measured data. Specifically, trained by the time series of a handful of perturbation events, we demonstrate that the trained machine can predict accurately whether the system will return to the functional state in response to a large, random perturbation, thereby reconstructing the attracting basin of the functional state. The working mechanism of the new machine is analyzed, and it is revealed that the success of the new machine is attributed to the good balance between the echo and fading properties of the reservoir network; the effect of noisy signals on the prediction performance is also investigated, and a stochastic-resonance-like phenomenon is observed. Finally, we demonstrate that the new technique can be also utilized to infer the attracting basins of coexisting attractors in typical chaotic systems.\\
%\begin{flushright} 
%{\bf Subject Areas:} Multistable Systems, Machine Learning 
%\end{flushright} 
%\noindent{\bf Subject Areas:} Multistability, Reservoir Computing, Power System, Attracting Basins
%\noindent{\bf Subject Areas:} Multistable Systems, Machine Learning
\end{abstract}

\maketitle

\section{Introduction}

A distinct feature of many real-world systems is that the asymptotic states the systems are settled into are dependent on their initial conditions. This feature, known as multistability in nonlinear science, has important implications to the functionality and security of a wide variety of complex systems, ranging from the human brain to ecosystems and power grids~\cite{Pisarchik2014,DD2016,Brain:2004,Ecosys:2011,Powergrid:2017}. For systems such as ecosystems and power grids, a critical challenge faced in system management is to estimate the possible influences that a non-small disturbance might induce, saying, for instance, whether a fire or illegal logging will switch a rainforest from the fertile state to the barren state~\cite{Ecosys:2011}; whether the failure of a power station or a single transmission line will trigger a large-scale blackout in a power grid~\cite{Powergrid:2017}. The concern of multistability is particularly important for today's world, as evidences indicate that many natural and man-made complex systems are currently working in the vicinity of their tipping points~\cite{Powergrid:2017,Tipping:FJF,TippingClimate:2023,Menck2013}. One approach to coping with the challenge is finding the attracting basins of the asymptotic states that the systems will settle into, namely the basins of attraction~\cite{BOA:1996}. Briefly, the attracting basin of an asymptotic state is the set of initial conditions which evolve into this state eventually. For nonlinear systems, the attracting basins are normally entangled with each other in a complicated fashion, e.g., the boundaries separating the basins are fractal or riddled~\cite{Book:Ott,TransientChaos}, while the size or volume of each basin determines the stability of the associated state~\cite{BOA:1996,Menck2013}. In particular, nearby the boundary of the basins, a small perturbation may lead the system to a completely different state. As such, to anticipate whether the disturbed system will return to its functional state, the key is to identify precisely the basin boundaries~\cite{BasinStabPower:CM2017,BasinStabPower:JP2018,BasinStabPower:YF2019,BasinStabPower:HK2019}. Though it is well recognized in nonlinear science that the boundaries of the attracting basins are defined by the stable manifolds of the nonattracting states, an accurate characterization of the boundaries in the global phase space remains a challenge, especially for high-dimensional nonlinear systems in which the number of nonattracting states is huge and the stable manifolds of these states are very complicated~\cite{TransientChaos,BasinZYZ2021}. Additionally, in realistic situations, the equations governing the system dynamics are usually unknown and what is available are only measured data, e.g., the time series measured from the transient dynamics of a power system, which renders the development of model-free, data-driven techniques an urgent task in analyzing the basin stability of real-world complex systems~\cite{BasinStabPower:CM2017,BasinStabPower:JP2018,BasinStabPower:YF2019,BasinStabPower:HK2019,Che2021,Datseris2022,TransientChaos,BasinZYZ2021}.

Recent advances in machine learning provide a bunch of new tools for the model-free inference of complex dynamical systems~\cite{ML:RevRMP2019,RC:Tang2020}. In particular, a special type of recurrent neural network named reservoir computer (RC) has been employed recently in the literature as a data-driven technique for inferring chaotic systems~\cite{RC:Maass2002,Jaeger2004}. From the perspective of dynamical systems, an RC can be regarded as a complex network of coupled nonlinear units which, driven by the input signals, generates the outputs through a readout function. In the simplest form, the implementation of RC consists of two phases, the training and predicting phases. In the training phase, the machine is fed with the time series measured from the target system, and the purpose of the training is to find the set of coefficients in the readout function for the best fitting of the training data. In the predicting phase, the input signals are replaced by the outputs and the machine is running as an autonomous system with fixed parameters, with the outputs giving the predictions. Although structurally simple, RC has demonstrated its superpower in a variety of applications, e.g., predicting the short-term state evolution of chaotic systems~\cite{Jaeger2004}, inferring the unmeasured variables~\cite{RC:Lu2017}, anticipating the long-term statistical properties of chaotic attractors~\cite{Pathak2017}, and replicating the dynamics of Hamiltonian chaos~\cite{HanZhang2021}. By a parallel architecture, RC has been also applied successfully to predict the state evolution of spatiotemporal systems~\cite{RC:Pathak2018,RC:Parlitz2018,RC:ParallelMachinePRL2022}. Besides predicting the dynamics of the target systems from which the training data are measured, recent studies also show that by introducing a parameter-control channel, RC is also able to infer the dynamics of some new systems it has never seen, e.g., based on the time series of a few exampling states, an RC is able to reconstruct the whole bifurcation diagram of a chaotic system~\cite{KLW2021,RC:Kim2021,HWFan2021}.  

Inspired by the multifunctionality of biological neural networks, attempts have been made recently on the learning of multiple attractors by a single RC~\cite{ZLu2020,Rohm2021,Flynn2021}. By the mechanism of invertible generalized synchronization, Lu {\it et al.} proposed a multifunctional learning framework in which an RC ``replicates" the dynamics of multiple attractors in the training phase~\cite{ZLu2020}, and later each attractor is ``retrieved" successfully by driving the RC with a transient time series in the predicting phase. R\"{o}hm {\it et al.} further demonstrated that, given that the driving series is sufficiently long, the RC can be trained by even a single noisy trajectory, while the trained machine is able to infer the coexisting attractors ``unseen" in the training data~\cite{Rohm2021}. An alternative framework of multifunctional RC is proposed by Flynn {\it et al.}~\cite{Flynn2021}, in which a ``blending technique" is employed to generate from the time series of two coexisting attractors the training data. It is shown that, by resetting the initial conditions of the reservoir network, the same machine is able to reproduce the trajectories of both attractors in the predicting phase. (As the initial conditions are set as the final state of the reservoir in the training phase, here a long time series of the target system is needed to drive the machine before making the predictions.) The fact that a single RC can replicate the dynamics of multiple attractors makes RC a potential solution to the model-free inference of the attracting basins in complex dynamical systems, as demonstrated preliminarily in the recent studies~\cite{Roy2022,Gauthier2022}. By the architecture of parameter-aware RC~\cite{KLW2021,RC:Kim2021,HWFan2021}, Roy {\it et al.} showed that the machine trained by the time series of a few exampling states is able to reconstruct the attracting basins of a new state with a reasonable precision~\cite{Roy2022}. Similar to the studies in Refs.~\cite{ZLu2020,Rohm2021,Flynn2021}, before making the predictions, a long time series from the desired attractor is also needed to drive the reservoir out of the transient. As the driving time series used to ``warm up" the machine is sufficiently long in these studies (which normally contains thousands of data points and sustains hundreds of system oscillations), the system will have already approached the asymptotic attractor during this time period, rendering the prediction of attracting basins by machine-learning techniques unnecessary~\cite{YZhang2022}. 

To cope with the problem of long driving series for initiating the RC, a new technique named next-generation RC (NGRC) has been employed recently for inferring the attracting basins of coexisting attractors in chaotic sytems~\cite{Gauthier2021,Gauthier2022}. Different from the conventional RC techniques in which a nonlinear reservoir and a linear output layer are adopted, NGRC utilizes a linear reservoir and a nonlinear output layer. Compared to the conventional RCs, NGRC is featured by employing fewer hyperparameters and using a much shorter ``warming up" time series. The latter makes NGRC a promising technique for inferring the attracting basins of complex dynamical systems, as demonstrated by Gauthier {\it et al.} in a recent study~\cite{Gauthier2022}. Specifically, it is shown in Ref.~\cite{Gauthier2022} that NGRC can be trained by the time series measured from a single attractor, while the trained machine is able to predict the fractal basins of the coexisting attractors in chaotic systems (e.g., the Li-Sprott system). Remarkably, compared to the standard RC, the ``warming up" series of NGRC is shortened by about $3$ orders of magnitude, while the prediction performance is improved by about $2$ orders of magnitude. However, as pointed out by Zhang {\it et al.} in Ref.~\cite{YZhang2022}, the excellent performance of NGRC comes at the cost of a priori knowledge of the nonlinearity of the target system. Specifically, a small uncertainty of the nonlinear functions of the target system might lead to a complete failure of the NGRC technique. This creates a catch-22 for NGRC: one has to balance between the prediction performance and the available system information (the nonlinearity of the target system) but can not have both~\cite{YZhang2022}. 

Noticing that the conventional RCs have their own catch-22 (i.e., no prior knowledge of the target dynamics is required but a long time series is needed to ``warm up" the reservoir before making the predictions)~\cite{ZLu2020,Rohm2021,Flynn2021,Roy2022}, it is reasonable to conjecture that the key for an RC to infer the dynamics of multistable systems might lie in the balance between the prediction performance and the prediction cost. For the conventional RCs, the cost is reflected in the length of the driving (``warming up") series; for the NGRC, the cost is reflected in the knowledge of the system dynamics. With this in mind, we propose in the present work a new technique of RC by balancing the prediction performance and the prediction cost, and utilizing the new RC to infer the attracting basins of a typical power system. We are able to show that, trained by the transient series of a handful of perturbation events, the machine is able to anticipate accurately whether the power system will return to the functional state when a non-small random disturbance is encountered given that the initial response of the system is measured, thereby reconstructing the attracting basin of the functional state in the phase space. We shall discuss in detail how the contradiction between prediction performance and cost is balanced by a new objective function, and demonstrate how the introduction of moderate noise can improve the prediction performance. Finally, we shall demonstrate that the new technique can be also utilized to infer the attracting basins of coexisting attractors in typical chaotic systems, signifying the capability of this new technique in learning the dynamics of general multistable systems. 

\section{Multistability of power system}

The power system studied in the present work is the voltage source converter (VSC) -- a key electronic device that has been widely adopted in renewable-energy-dominated power systems for maintaining the synchronization relationship between the generators (e.g., the wind farms and photovoltaic plants) and the power grid through the technique of phase-locked loop (PLL)~\cite{PowerSys:Book}. The failure of VSC leads to generally the dysfunction or collapse of power generators, which, according to industrial reports~\cite{VSCFault-1}, contributes to a significant portion of the accidents in modern power systems. While sophisticated models have been proposed in the literature for the dynamics of PLL-based VSC, here we adopt the generalized swing model proposed in Ref.~\cite{RMa2022} for simplicity and demonstration purposes. The model reads
\begin{equation}
\begin{cases}
\dot{\theta}=\omega,\\
\dot{\omega}=I-\sin{\theta}-(\alpha\cos{\theta}-D)\omega,
\end{cases}
\label{model}
\end{equation}
where $\theta$ and $\omega$ denote, respectively, the phase and frequency mismatches between the generator and the grid, $I$ represents the dimensionless input power (determined by the operating conditions of the generator), $D$ stands for the constant equivalent damping coefficient (determined by both the operating conditions and the integral coefficients of the PLL module), and $\alpha$ characterizes the state-dependent equivalent damping coefficient (determined by the ratio of the proportional and integral coefficients of the PLL module). The three parameters $(D, I, \alpha)$ are positively defined on the physical ground, and Eq.~(\ref{model}) describes essentially the stability of the synchronization state of the power generator~\cite{RMa2022}. Different from the classical swing models in which the damping coefficient is a constant~\cite{PowerSys:Book}, the generalized swing model is featured by the state-dependent damping coefficient, $\hat{D}(t)=\alpha\cos{\theta}-D$. This feature makes the stability of the renewable-energy-dominated power systems significantly different from that of synchronous-generator-dominated power systems, and raises a series of new challenges in system analysis and management, e.g., the attracting basin of the functional state shows a fish-like pattern~\cite{BS:2014,PSys:ZZ2021,PSys:CZ2020}. 

\begin{figure}[tbp]
\begin{center}
\includegraphics[width=0.75\linewidth]{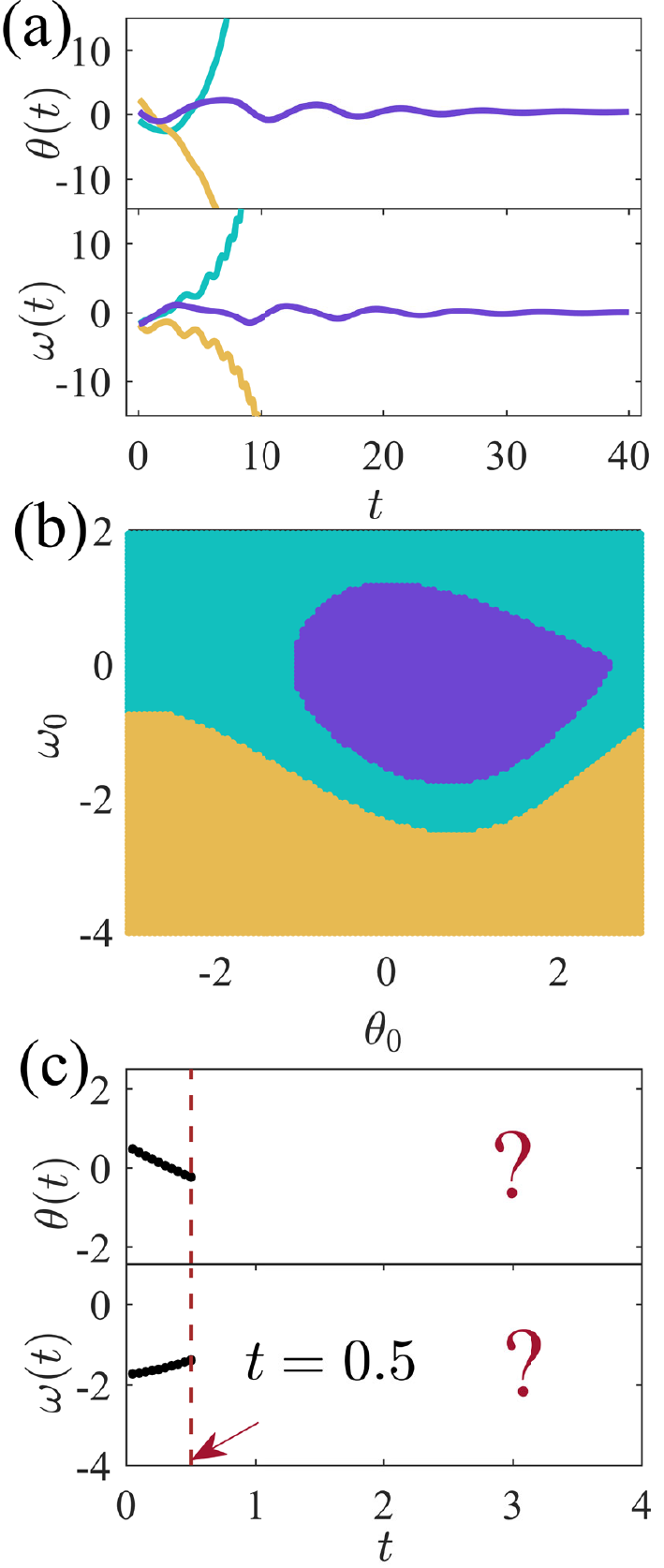}
\centering\caption{Multistability of the power system described by the generalized swing model. (a) The time evolution of the system started from typical initial conditions. Purple curve: the stable case in which $\omega$ is damped to $0$ after a transient period about $T=30$. Green curve: the system is unstable. $\omega$ diverges to $+\infty$ with time. Yellow curve: the system is unstable. $\omega$ diverges to $-\infty$ with time. (b) Basins of attraction obtained by the approach of model simulation. Purple region: the attracting basin of the operating state $\omega=0$. Green region: the attracting basin of the diverging state $\omega=+\infty$. Yellow region: the attracting basin of the diverging state $\omega=-\infty$. (c) The initial response of the system to a large, random perturbation, which contains $10$ data points. This short time series will be used as the guiding series for the machine to anticipate the asymptotic state.} 
\label{fig1}
\end{center}
\end{figure}

To illustrate the complex dynamics of the generalized swing model, we set $(I,D,\alpha)=(0.4,0.39,0.7)$ in Eq.~(\ref{model}) and, by the approach of model simulations, plot in Fig.~\ref{fig1}(a) the system evolutions started from several initial conditions $(\theta_0,\omega_0)$. In numerical simulations, Eq.~(\ref{model}) is solved by the 4th-order Runge-Kutta algorithm with the time step $\delta t=0.05$. Since $\theta$ and $\omega$ represent, respectively, the phase and frequency mismatches between the generator and the grid, different initial conditions can be regarded as different disturbances added onto the power system. The system is considered as stable if $\omega$ is damped to $0$ after a short transient and as unstable if $\omega$ is diverging with time. We see in Fig.~\ref{fig1}(a) that, though the initial conditions (disturbances) are very close (similar) to each other, the asymptotic states the power system is settled into are completely different: the system is returned to the functional (operating) state with $\omega=0$ for one of the initial conditions (the purple curve), but is diverged to infinity for the other two initial conditions (the green and yellow curves). In particular, in the stable case, the system is returned to the operating state ($\omega=0$) after a short transient period $T\approx 30$, i.e., the disturbance is disappeared after about $4$ oscillations. Treating $\omega=\pm\infty$ as two different attractors, we plot in Fig.~\ref{fig1}(b) the attracting basins of the three asymptotic states ($\omega=0$, $\infty$, and $-\infty$) in the phase space based on the results of model simulations. We see that the basins are separated from each other by irregular boundaries. (More complex boundaries can be generated by decreasing the parameter $D$, which will be discussed later.)  

Whereas Fig.~\ref{fig1}(b) serves as the reference manual for system management in dealing with any random disturbance, the construction of the attracting basins is extremely challenging for realistic systems. One reason is that the accurate model describing the power system is unavailable, which makes a brute-force search of the attracting basins by the approach of model simulation as did in plotting Fig.~\ref{fig1}(b) infeasible. Without an accurate model, any attempt for constructing the basins must be based on measured data. On the other hand, as large-perturbation events in power systems are rare and a complete record of the system response to large perturbations is difficult in practice (especially for incidents leading to system collapses), it is also impossible to reconstruct the basins by collecting data for each possible perturbation (initial condition). From the perspective of power-system management, what is available is just the time series measured from a few perturbation events, while the mission is to anticipate whether the system will return to the operating state in the presence of a random perturbation. In particular, by noticing the abnormal behavior of a power system for a very short episode (which shows no sign of the asymptotic state), the system operator needs to decide promptly whether actions should be taken to prevent the system from collapsing (if it is unstable) or just leave the system alone (if it is stable)~\cite{PowerSys:Book}. For the generalized swing model studied here, the question can be rephrased as follows: given that the time series of a handful of perturbation events are measured and the data are available, can we predict, based on the initial responses of the power system to a random perturbation, whether the system will restore the operating state after the transient? An example of the initial responses of the system to a random perturbation is plotted in Fig.~\ref{fig1}(c), which contains $l=10$ data points and sustains for only a time period of $T=0.5$ (less than one-tenth of the system oscillation). Our main objective in the present work is to predict, based on the information of the initial responses, whether the power system will restore its functional state. We are going to demonstrate that this goal can be achieved by a new technique generalized from the standard RCs. In specific, we are able to show that by balancing the echo and fading properties of the reservoir network, the machine trained by the time series of a few perturbation events [as depicted in Fig.~\ref{fig1}(a)] is able to infer from the short time series of the initial responses whether the perturbed system will return to the operating state or collapse eventually, therefore reconstructing the attracting basins of the power system.

\section{Balanced reservoir computer}

The new RC technique proposed in the current study is similar to the conventional ones in architecture~\cite{RC:Lu2017,Pathak2017,HanZhang2021,RC:Pathak2018}, but is different in hyperparameter optimization and machine implementation. Like the conventional RCs, the new RC consists of also three modules: an input layer, a reservoir, and an output layer. The input layer is characterized by the matrix $\bm{W}_{in}\in\mathbb{R}^{n\times d}$, whose function is coupling the input vector $\bm{u}(t)\in\mathbb{R}^{d}$ to the reservoir. The elements of $\bm{W}_{in}$ are randomly drawn from a uniform distribution within the range $[-\sigma, \sigma]$. The reservoir is represented by a complex network of $n$ dynamical nodes, with the initial states of the nodes being randomly chosen from the interval $[-1,1]$. The state of the reservoir network, $\bm{r}(t)\in \mathbb{R}^{n}$, is updated as
\begin{equation}\label{rc1}
\bm{r}(t+\Delta t)=(1-\alpha)\bm{r}(t)+\alpha\tanh[\bm {A}\bm{r}(t)+\bm{W}_{in}\bm{u}(t)].
\end{equation}
Here, $\Delta t$ is the time step for updating the reservoir, $\alpha$ is the leaking coefficient, $\tanh$ is the hyperbolic tangent function, and $\bm{A}\in \mathbb{R}^{n\times n}$ is a weighted adjacency matrix characterizing the coupling relationship between nodes in the reservoir. The adjacency matrix $\bm{A}$ is constructed as a sparse random Erd\"{o}s-R\'{e}nyi network: with the probability $p$, each element of the matrix is arranged a nonzero value drawn randomly from the interval $[-1,1]$. The matrix $\bm{A}$ is rescaled to make its spectral radius equal $\lambda$. The output layer is characterized by the matrix $\bm{W}_{out}\in \mathbb{R}^{d\times n}$, which generates the output vector, $\bm{v}(t)\in \mathbb{R}^{d}$, according to the equation 
\begin{equation}\label{rc2}
\bm{v}(t+\Delta t)=\bm{W}_{out}\bm{\tilde{r}}(t+\Delta t),
\end{equation}
where $\bm{\tilde{r}}\in \mathbb{R}^{n}$ is the new state vector obtained from the reservoir state (i.e., $\tilde{r}_i=r_i$ for the odd nodes and $\tilde{r}_i=r_i^2$ for the even nodes)~\cite{RC:Pathak2018}, and $\bm{W}_{out}$ is the output matrix to be obtained through a training process. 

The implementation of the machine consists of three phases (training, validation, and prediction), and each phase contains two stages (listening and working). In the training phase, the reservoir is driven by the time series of the target system first for a transient period of $l$ steps (the listening stage) and then for a long period of $L$ steps (the working stage). The purpose of the listening stage is to make the reservoir network ``forget" its initial conditions, so as to realize the ``echo" property of the machine in the working stage~\cite{Jaeger2004}. The mission of the working stage in the training phase is to find a suitable output matrix $\bm{W}_{out}$, so that the output vector $\bm{v}(t+\Delta t)$ as calculated by Eq. (\ref{rc2}) is as close as possible to the input vector $\bm{u}(t+\Delta t)$ for the data points at $t=(l+1)\Delta t\ldots,(l+L)\Delta t$ in the training series. This can be done by minimizing the cost function with respect to $\bm{W}_{out}$~\cite{RC:Lu2017,Pathak2017,HanZhang2021,RC:Pathak2018}, which gives
\begin{equation}\label{rc3}
\bm{W}_{out}=\bm{U}\bm{V}^T(\bm{V}\bm{V}^T+\eta\mathbb{I})^{-1}.
\end{equation}
Here, $\bm{V}\in \mathbb{R}^{n\times L}$ is the state matrix whose $k$th column is $\bm{\tilde{r}}[(l+k)\Delta t]$, $\bm{U}\in \mathbb{R}^{d\times L}$ is a matrix whose $k$th column is $\bm{u}[(l+k)\Delta t]$, $\mathbb{I}$ is the identity matrix, and $\eta$ is the ridge regression parameter for avoiding the overfitting. The hyperparameters of the machine include $n$ (the size of the reservoir network), $p$ (the connectivity of the reservoir network), $\lambda$ (the spectral radius of the reservoir network), $\sigma$ (the range defining the input matrix), $\alpha$ (the leaking coefficient), and $\eta$ (the ridge regression parameter). All these hyperparameters are fixed at the construction of the machine, and each set of hyperparameters generates a unique output matrix. Once the output matrix is obtained, the training phase is completed. The set of hyperparameters and the output matrix together define a specific machine, which will be used later for validation and prediction purposes.

The machine which gives a good performance on the training data might not perform well on the validating (testing) data. The finding of the optimal set of hyperparameters performing well for both the training and testing data is the mission of the validation phase. For the conventional RCs, the performance of the machine is generally evaluated by the length of the time series that is accurately predicted, and the optimal machine is identified as the one giving the longest prediction length (horizon). To achieve a longer prediction, a general approach is to prolong the memory of the reservoir, e.g., setting its dynamics at the edge of chaos through modifying the structure of the reservoir network~\cite{Langton1990,RC:NB2004,RC:LYC2019,Carroll:2022} (counterexamples are given in Ref.~\cite{Carroll2020}). While prolonged memory improves the ``echo" property of the reservoir and thereby extends the prediction horizon, it also slows down the convergence of the reservoir dynamics in responses to external drivings. To emulate the dynamics of a chaotic system, a necessary condition is the establishment of the (generalized) synchronization relationship between the target system and the reservoir~\cite{ZLu2020}, with the synchronization speed characterized by the largest conditional Lyapunov exponent of the reservoir network. For a multifunctional RC capable of replicating multiple attractors, the synchronization speed determines the length of the time series required for driving the reservoir to the desired attractor: the larger the synchronization speed, the shorter the driving series. From the perspective of basin inference, this means that, to anticipate the asymptotic state of the target system based on its initial dynamics, the reservoir should be designed with a short memory. This requirement, however, is contradictory to the requirement posed for predicting the system evolution (in which a longer reservoir memory is desired). Therefore, to infer the attracting basins of multiple systems, a balance must be made between the ``echo" (state-evolution prediction) and ``fading" (synchronization) properties of the reservoir. 

\begin{figure}[tbp]
\begin{center}
\includegraphics[width=0.9\linewidth]{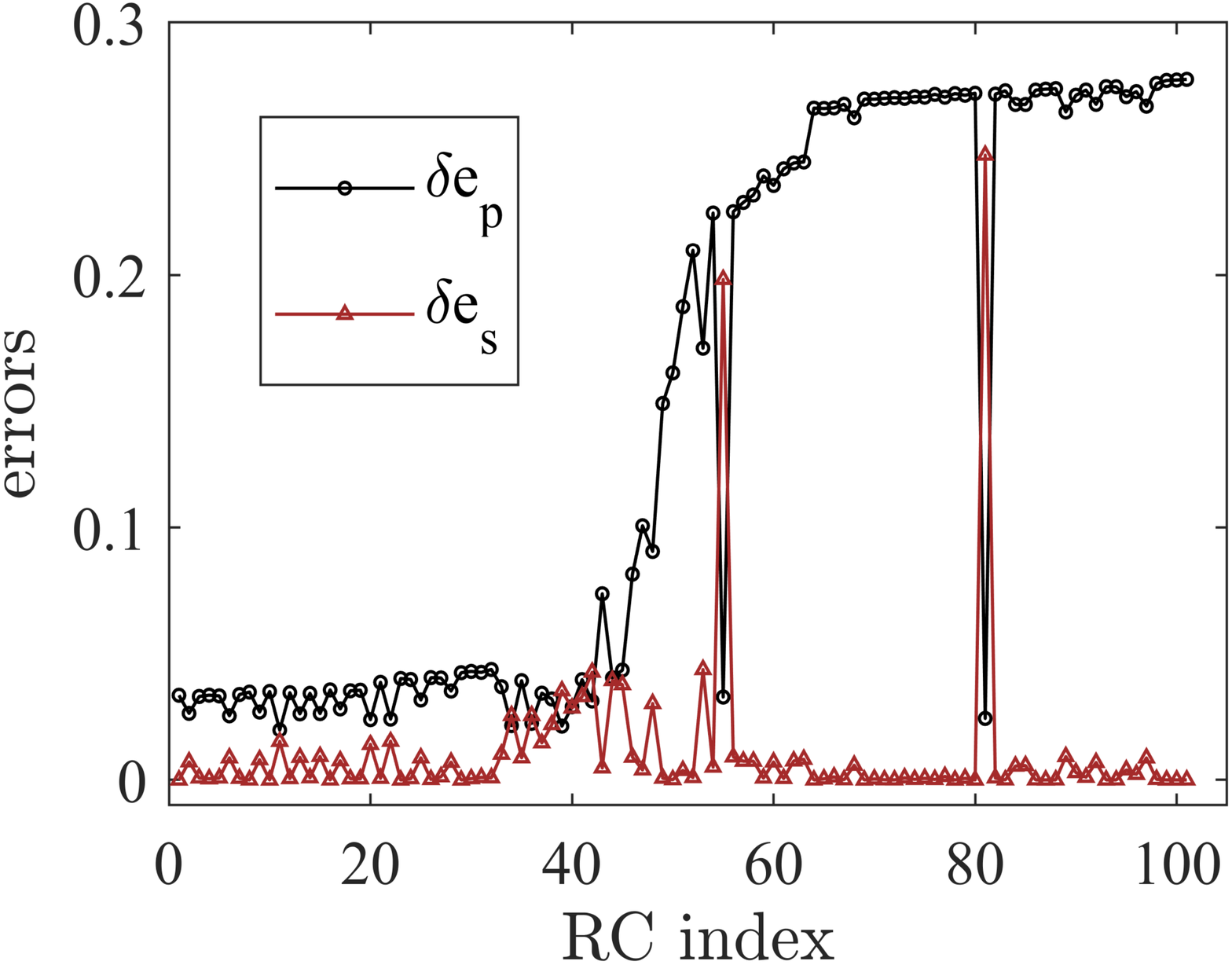}
\centering\caption{The contradictory requirements of state-evolution prediction and reservoir-network synchronization in designing multifunctional RC. Shown is the performance of $100$ RCs on the tasks of state-evolution prediction and reservoir-network synchronization. $\delta e_p$ denotes the error of state-evolution prediction (black circles). $\delta e_s$ denotes the error of reservoir synchronization (red triangles). The RCs are indexed by the increasing order of $\delta e$ under the balancing parameter $\beta=10$. Anti-correlation is observed between $\delta e_p$ and $\delta e_s$.} 
\label{fig2}
\end{center}
\end{figure}

We take the power system introduced in Sec. II as the example to demonstrate the contradictory requirements of state-evolution prediction and reservoir synchronization in designing multifunctional RC. Shown in Fig.~\ref{fig2} is the performance of $100$ RCs on the tasks of state-evolution prediction and reservoir-network synchronization. The RCs are identical in network size (the hyperparameter $n$ is identical while the network structures are generated independently), but are different in the other hyperparameters, $(p,\lambda,\sigma,\alpha,\eta)$. The performance of state-evolution prediction is evaluated by the error $\delta e_{p} =\left<\Vert\bm{u}(t)-\bm{v}(t)\Vert\right>$, with $\Vert\cdot\Vert$ the $\mathcal{L}^2$-norm and $\left<\cdot\right>$ the time-average function. (In predicting the system evolution, the machine is first driven by a short time series for $l=10$ steps in the open-loop configuration, then the machine is closed and operating in the closed-loop configuration for $L=1000$ steps.) Here $\bm{u}(t)$ and $\bm{v}(t)$ represent, respectively, the ground-truth results and the predicted results. The synchronization performance of the reservoir network is measured by the error $\delta e_{s} =\Vert \bm{r}_{\tau}-\bm{r}'_{\tau} \Vert$. Here, $\bm{r}_{\tau}$ is the instant state of the reservoir network after being driven by the input series for $\tau=10$ steps, and $\bm{r}'_{\tau}$ is the state generated by the same reservoir network but started from a different initial state. Clearly, the smaller the error $\delta e_p$, the better the prediction on system evolution; the smaller the error $\delta e_s$, the faster the establishment of generalized synchronization between the reservoir and the target system~\cite{GS1996}. While an ideal RC is expected to enjoy both a smaller prediction error and a smaller synchronization error, the results in Fig.~\ref{fig2} imply that such an ideal RC is impossible. To be specific, we see in Fig.~\ref{fig2} that the two errors, $\delta e_p$ and $\delta e_s$, are {\it anti-correlated}, i.e., the machine with a smaller $\delta e_p$ also possesses a larger $\delta e_s$, and vice versa.  

For the contradictory requirements revealed above, we propose to evaluate the performance of multifunctional RC by the error (objective function)
\begin{equation}
\delta e=\delta e_p+\beta\delta e_s,
\label{error}
\end{equation}
with $\beta$ the parameter balancing the performances of state-evolution prediction ($\delta e_p$) and reservoir synchronization ($\delta e_s$). In applications, the value of $\beta$ is chosen in such a way that the two terms on the RHS of Eq.~(\ref{error}) are of the same order, i.e., $\delta e_p=\mathcal{O}(\beta\delta e_s)$. As $\delta e_s$ is dependent on $\tau$ (the length of the driving series for synchronizing the reservoir) and $\delta e_s\ll \delta e_p$ in general, we have $\beta\gg 1$ in most cases. Adopting $\delta e$ as the new objective function, the next step is to find the optimal set of hyperparameters giving the best overall performance for the testing data, namely the validation phase. This is done by scanning each hyperparameter over a certain range through the conventional optimization algorithms such as the Bayesian and surrogate optimization algorithms~\cite{KLW2021}. As the optimal machine such constructed takes into account both the prediction and synchronization performances, we name the new machine the balanced RC. Finally, in the predicting phase, with the reservoir network and the hyperparameters being fixed, we first drive the machine by the guiding time series measured from the initial responses of the system [as depicted in Fig.~\ref{fig1}(c)] in the open-loop configuration (i.e., the listening stage). Then, we operate the machine in the closed-loop configuration (i.e., the working stage) and, based on the outputs, anticipate the asymptotic state that the system will settle into eventually.  

\section{Results}

We proceed to utilize the technique of balanced RC to infer the asymptotic state and the attracting basins of multistable dynamical systems, including the power system introduced in Sec. II and two typical chaotic systems.

\subsection{Inferring the attracting basins of power system}

\begin{figure}[tbp]
\begin{center}
\includegraphics[width=0.75\linewidth]{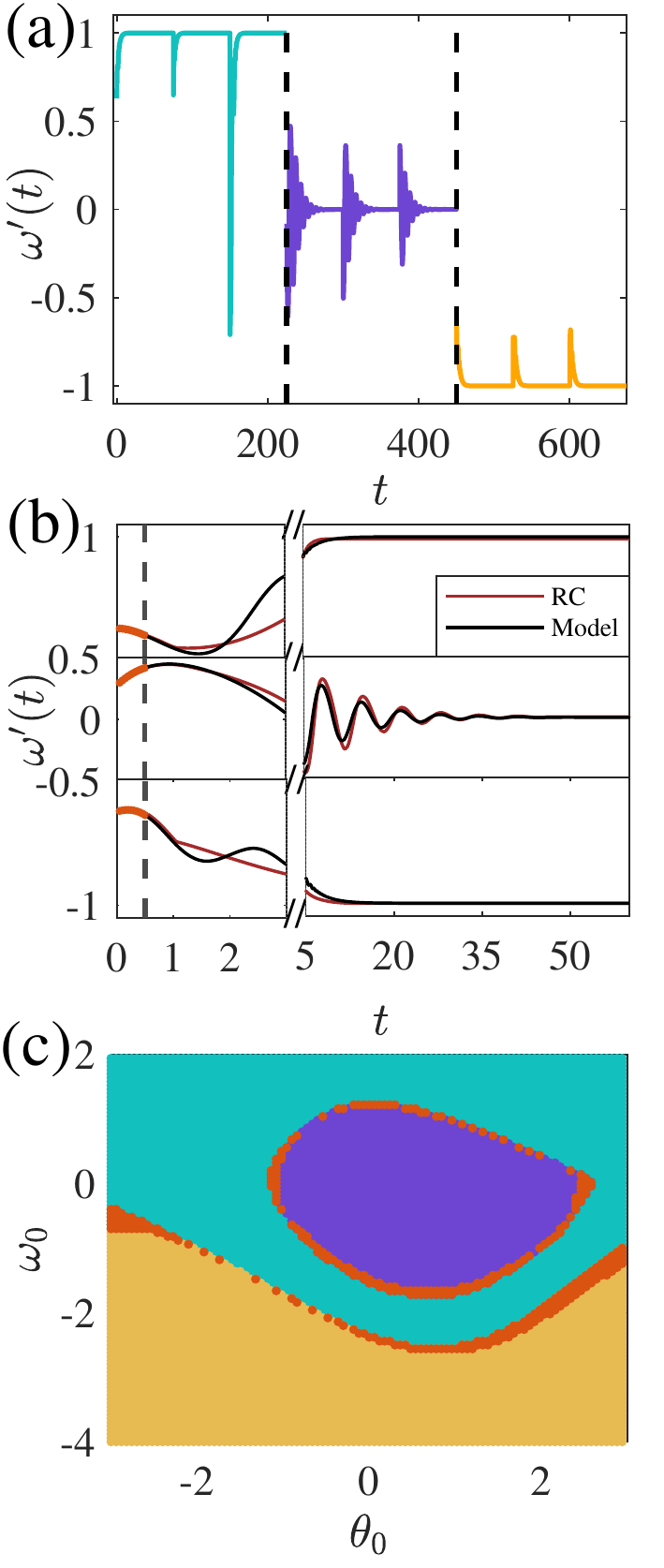}
\centering\caption{Inferring the attracting basins of the power system. (a) The time series of the normalized variable $\omega'(t)$ in the training data. The time series is composed of $N=9$ segments, and each segment contains $\hat{L}=1500$ data points.  (b) Typical predictions made by the machine on the state evolution of the perturbed system. Black curves are the results obtained by model simulations. Red curves are the results predicted by the machine. (c) The attracting basins of the power system predicted by the machine. Initial conditions that are attracted to the operating state ($\omega'=0$), the positive-diverging state ($\omega'=1$) and the negative-diverging state ($\omega'=-1$) are represented by purple, green and yellow points, respectively. Red points denote the false predictions, which are distributed along the basin boundaries. The prediction accuracy is about $96.7\%$.} 
\label{fig3}
\end{center}
\end{figure}

We start by preparing the training and testing datasets. Setting the parameters $(I,D,\alpha)=(0.4,0.39,0.7)$ in Eq.~(\ref{model}), we obtain by numerical simulation the time evolution of the system state started from different initial conditions. The initial conditions of the system are randomly chosen from the ranges $\theta_0\in[-3, 3]$ and $\omega_0\in [-4, 2]$, and each time series contains $\hat{L}=1500$ data points (sustains for $T=\hat{L}\delta t=75$ time units). According to the asymptotic state that the system is settled into ($\omega\in\{0,+\infty,-\infty\}$), we divide the time series into $3$ groups. We pick out $m=3$ time series from each group, and $N=9$ time series are selected in total. We then normalize the data by the arctangent function: $\theta'=2\arctan{\theta}/\pi$ and $\omega'=2\arctan{\omega}/\pi$. The $N$ normalized time series form the training data. Plotted in Fig.~\ref{fig3}(a) is the time series of $\omega'$ in the training data, in which the segments are colored according to the asymptotic states. The testing data is prepared in exactly the same way ($m=3$ and $N=9$), except that the time series are generated by different initial conditions.

We next find the optimal machine for basin inference according to the new objective function defined by Eq.~(\ref{error}). For simplicity, we fix the size of the reservoir as $n=500$, while tunning only the hyperparameters $(p,\lambda,\sigma,\alpha,\eta)$ in searching the optimal machine. The ranges over which the hyparameters are searched are $p\in (0,1)$, $\lambda\in(0,3)$, $\sigma\in(0,3)$, $\alpha\in(0,1)$, and $\eta\in(1\times 10^{-10},1\times 10^{-2})$. The optimal hyperparameters are obtained after $300$ trials in the parameter space with the help of the ``optimoptions" function in Matlab. For each set of hyperparameters that define a machine, we first calculate the synchronization error $\delta e_s$ of the reservoir network based on the training data, and then evaluate the prediction error $\delta e_p$ based on the testing data. In calculating $\delta e_s$, the driving series contains $\tau=10$ data points, and the result is averaged over $50$ realizations. In calculating $\delta e_p$, the driving series contains $l=10$ data points, and each training (testing) series contains $L=1490$ data points. In this application, the balancing parameter is chosen as $\beta=10$, which is estimated according to the prediction and synchronization errors of a randomly selected machine. The validation phase ends up with the optimal hyperparameters $(p,\lambda,\sigma,\alpha,\eta)=(0.480,0.033,2.917,0.574,3.458\times 10^{-4})$, which, together with the associated output matrix, defines the optimal machine.

Is the machine designed and optimized in such a way capable of inferring the asymptotic state and the attracting basins of the power system? We first check the capability of the machine in predicting the asymptotic state when the initial responses of the power system to a random disturbance are measured [see Fig.~\ref{fig1}(c)]. In doing this, we first drive the machine by the guiding series (containing $l=10$ data points) in the open-loop configuration, then, using the final state of the reservoir as the initial state, operate the machine in the closed-loop configuration. Typical examples of the predictions are plotted in Fig.~\ref{fig3}(b). We see that, though the transient states of the system evolution are not precisely predicted, the machine does anticipate accurately the asymptotic state the power system settles into eventually. We proceed to check the capability of the machine in inferring the attracting basins of the operating state ($\omega'=0$), the positive-diverging state ($\omega'=+1$) and the negative-diverging state ($\omega'=-1$). The results predicted by the machine are plotted in Fig.~\ref{fig3}(c), in which the regions colored in purple, green and yellow represent, respectively, the attracting basins of the operating state, the positive-diverging state and the negative-diverging state. [In plotting Fig.~\ref{fig3}(c), the system is regarded as having settled into the operating state if $\left|\omega'\right|<1\times 10^{-2}$ after $1500$ iterations ($75$ time units), and is regarded as having settled into the positive-diverging (negative-diverging) state if $\omega'>0.99$ ($\omega'<-0.99$) after the same number of iterations]. Compared to the ground-truth results plotted in Fig.~\ref{fig1}(b), the machine predicts correctly the asymptotic state for $96.7\%$ of the initial conditions. The false predictions, which are marked by the red points in Fig.~\ref{fig3}(c), are distributed along the basin boundaries. A close look at the false predictions shows that in these cases the predicted trajectories are diverged from the true trajectories shortly after the machine is operating in the closed-loop configuration (not shown), signifying the sensitivity of the system dynamics nearby the basin boundaries~\cite{Book:Ott,TransientChaos}.

\subsection{Impacts of sampling series and noise}

\begin{figure}[tbp]
\begin{center}
\includegraphics[width=0.75\linewidth]{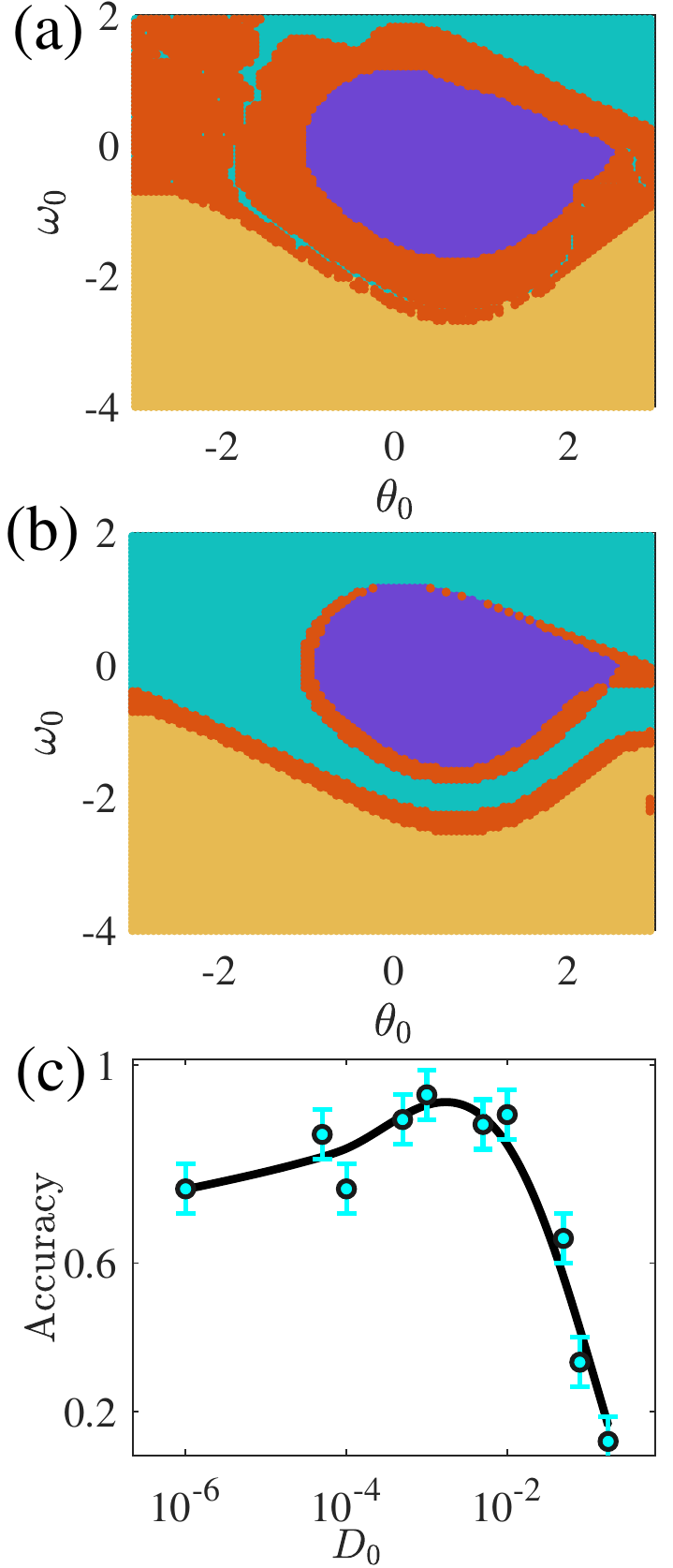}
\centering\caption{The impacts of sampling series and noise on the inference of attracting basins. (a) The attracting basins predicted by the machine trained by the tailored training data consisting of $N=6$ sampling series (each asymptotic state contributes $m=2$ time series). The prediction accuracy is about $70\%$. (b) The attracting basins predicted by the machine trained by noisy signals. The noise amplitude is $D_0=1\times 10^{-5}$. The prediction accuracy is about $85\%$.  (c) The phenomenon of stochastic resonance in basin prediction. Shown is the variation of the prediction accuracy with respect to the noise amplitude $D_0$. The prediction accuracy reaches its maximum at about $D_0=2\times 10^{-3}$. Results are averaged over $50$ realizations. Error bars denote the variances. False predictions are represented by red dots in (a) and (b).} 
\label{fig4}
\end{center}
\end{figure}

In learning the dynamics of multistable systems by the conventional RC techniques, a general requirement is that the training data should be composed by the time series of all the coexisting attractors~\cite{ZLu2020,Flynn2021}. This requirement also applies to the technique of balanced RC proposed in our present work. Additional simulations show that, if the training data is composed by the time series of only one or two asymptotic states, the accuracy of the prediction will be clearly decreased (not shown). Even though the training data contains the time series of all the attractors, the performance of the machine is still influenced by the number of the sampling series. To show an example, we delete one time series for each asymptotic state in the training data shown in Fig.~\ref{fig3}(a) (i.e. $m=2$ and $N=6$), and use the tailored data to optimize and train a new machine. The hyperparameters of the new machine are $(p,\lambda,\sigma,\alpha,\eta)=(0.804,0.852,2.690,0.965,1.552\times 10^{-4})$, and the balancing parameter is chosen as $\beta=15$. The attracting basins predicted by the new machine are plotted in Fig.~\ref{fig4}(a). Compared to the results plotted in Fig.~\ref{fig3}(c), we see in Fig.~\ref{fig4}(a) that a significant portion of the predictions are incorrect (the red points). The prediction accuracy of the new machine is about $70\%$, which is clearly smaller than that of the original machine ($\sim 96.7\%$). [The adoption of more sampling series, e.g., $m=4$ samplings for each asymptotic state, will improve the prediction performance, but the improvement is marginal (not shown).]    

In predicting real-world dynamical systems by machine-learning techniques, an important concern is the possible impacts of noise~\cite{Noise:CMB1995,Noise:KCJ1996}. Compared to noise-free signals, the machine performance is normally deteriorated when noise is introduced~\cite{Noise:MS2021,Noise:Nathe2023}, e.g., the prediction horizon is shortened and the replicated dynamics is unstable. Counterintuitive examples have been reported recently in predicting chaos using RC, in which it is found that under some circumstances the introduction of noise could improve the prediction performance~\cite{Noise:Estebanez2019,Noise:SR2023}. To be specific, it is demonstrated in Ref.~\cite{Noise:SR2023} that, with the increase of the noise amplitude, the prediction performance of RC is firstly increased and then decreased, showing the typical phenomenon of stochastic resonance in nonlinear science~\cite{SR}. Inspired by these studies, we investigate here the impacts of noise on the prediction performance of balanced RC. To introduce noise perturbations, we modify the dynamical functions of the power system to
\begin{equation}
\begin{cases}
\dot{\theta}=\omega,\\
\dot{\omega}=I-\sin{\theta}-(\alpha\cos{\theta}-D)\omega+D_0\xi(t),
\end{cases}
\label{modelnoise}
\end{equation}
with $\xi(t)$ the Gaussian white noise (zero mean and unity variance) and $D_0$ the noise amplitude. Setting $D_0=1\times 10^{-5}$, we generate the training data by simulating Eq.~(\ref{modelnoise}) numerically and then use the noisy signals to train and optimize a new machine. As did for the noise-free case [see Fig.~\ref{fig4}(a)], we collect for each asymptotic state $m=2$ sampling series, and $N=6$ sampling series are collected in total. The hyperparameters of the new machine are $(p,\lambda,\sigma,\alpha,\eta)=(0.404,0.752,2.637,0.738,8.341\times 10^{-4})$, and the balancing parameter is chosen as $\beta=12$. The attracting basins predicted by the new machine are plotted in Fig.~\ref{fig4}(b). We see that, compared with the noise-free case, the prediction accuracy is significantly improved for the noisy signals ($\sim 85\%$). To have a global picture on the impacts of noise on the prediction performance, we plot in Fig.~\ref{fig4}(c) the variation of the prediction accuracy with respect to the noise amplitude. The ``bell shape" variation shows clearly the occurrence of stochastic resonance at the optimal noise amplitude $D_0\approx 2\times 10^{-3}$. 

We note that the underlying mechanism of the stochastic-resonance phenomenon observed here is different from the one revealed in Ref.~\cite{Noise:SR2023}. In Ref.~\cite{Noise:SR2023}, noise is added onto the measured data directly (i.e. the measurement noise), and the emergence of stochastic resonance is attributed to the competition between the performances of the machine on the training and testing data. In such a situation, noise plays essentially the same role as the ridge regression parameter ($\eta$): avoiding overfitting in machine training. Different from that, in our current study noise is added onto the system dynamics (i.e. the intrinsic noise), and stochastic resonance emerges as a competition between the extended transient dynamics and the data quality. To be specific, by increasing the noise amplitude, the transient period for the perturbed power system to be settled into the asymptotic state will be extended, which provides more information about the system dynamics and therefore is beneficial to the learning. However, if the noise is too strong, the quality of the data will be reduced, making the replication of the system dynamics inaccurate. (A similar phenomenon is also observed in Ref.~\cite{HWFan2022}, in which it is shown that the introduction of intrinsic noise could extend the transient time for synchronization, which improves the performance of the RC in predicting the critical point for synchronization.)

\subsection{Inferring complicated attracting basins}

\begin{figure}[tbp]
\begin{center}
\includegraphics[width=0.75\linewidth]{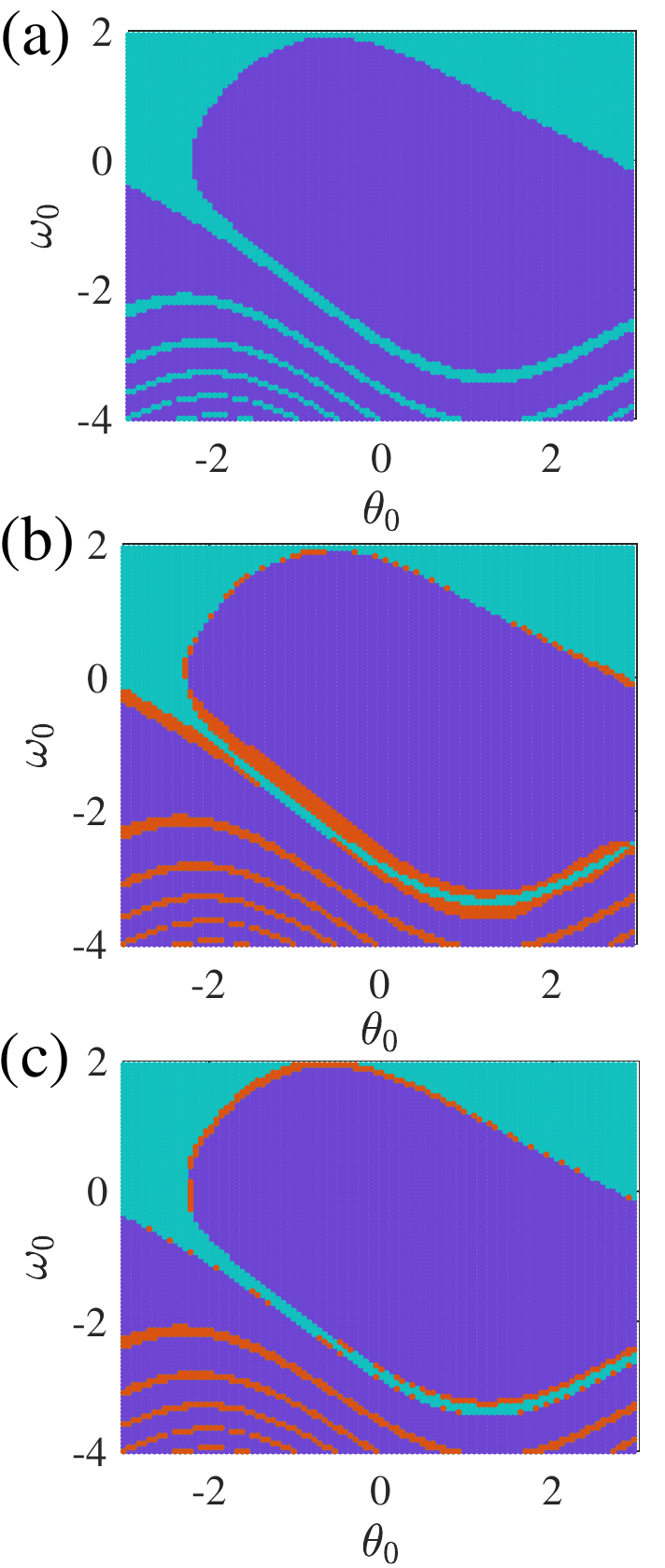}
\centering\caption{The attracting basins of the power system for the parameter $D=0.06$. (a) The results obtained by model simulations. (b) The results predicted by the machine based on noise-free signals. The prediction accuracy is about $91\%$. (c) The results predicted by machine based on noisy signals with noise amplitude $D_0=1\times 10^{-2}$. The prediction accuracy is about $96\%$. Purple region: the attracting basin of the operating state. Green region:  the attracting basin of the diverging state. Red dots: the false predictions.} 
\label{fig5}
\end{center}
\end{figure}

How about complicated attracting basins? By changing the parameter $D$, more complicated basins can be observed in the power system~\cite{RMa2022}. Setting $D=0.06$ in Eq.~(\ref{model}), we plot in Fig.~\ref{fig5}(a) the attracting basins of the operating ($\omega=0$) and diverging ($\omega=+\infty$) states in the phase space. Compared to the basins plotted in Fig.~\ref{fig1}(b) ($D=0.39$), we see that the basins are more complicated. In particular, the attracting basin of the operating state is composed of several disconnected regions, showing the feature of the fish-like pattern~\cite{RMa2022}. We next utilize the technique of balanced RC to infer the fish-like basins.

For the generalized swing model, the transient period for the perturbed system to be settled into the asymptotic states is decreased with the parameter $D$. For the parameter $D=0.06$, the system is damped to the operating state after only one or two oscillations. To obtain sufficient information for training (validating) the machine, here we collect from each asymptotic state $m=5$ time series, while each time series contains $\hat{L}=1000$ data points. As there are only two asymptotic states ($\omega=0$ or $+\infty$) in the case, the training (testing) data contains $N=10$ time series and $10000$ data points in total (which is longer than the training data used in Fig~\ref{fig3}). The listening (guiding) series contains $l=10$ data points, and the time series predicted by the machine contains $2000$ data points. The balancing parameter is chosen as $\beta=30$, and the optimal hyperparameters are $(p,\lambda,\sigma,\alpha,\eta)=(0.758,0.046,1.689,0.586,6.91\times 10^{-5})$. The criteria for inferring the asymptotic state are the same as in Fig.~\ref{fig3}(c). The results predicted by the machine are plotted in Fig.~\ref{fig5}(b). The prediction accuracy is about $91\%$. We see that, compared to the results of $D=0.39$ [see Fig.~\ref{fig3}(c)], the prediction accuracy is decreased for the fish-like attracting basins (even with more training data). 

Introducing noise to the system dynamics, we check further the impact of noise on the prediction performance. Setting the noise amplitude as $D_0=1\times 10^{-2}$, we regenerate the training and testing data by simulating Eq.~(\ref{modelnoise}), and then obtain a new machine by the same procedures as did for the noise-free case. The new set of hyperparameters are $(p,\lambda,\sigma,\alpha,\eta)=(0.854,0.086,2.401,0.489,9.161\times 10^{-5})$ and the balancing parameter is still $\beta=30$. The results predicted by the machine are plotted in Fig.~\ref{fig5}(c). The prediction accuracy is about $96\%$. Again, we see that the introduction of noise can improve the prediction performance, which is in consistent with the results obtained for the parameter $D=0.39$ [see Fig.~\ref{fig4}(c)].

\subsection{Inferring the attracting basins of coexisting attractors in chaotic systems} 

We finally utilize the proposed technique to infer the attracting basins of coexisting attractors in typical chaotic systems. Before presenting the detailed results, we would like to mention briefly some challenges met currently in the literature in predicting the attracting basins of chaotic systems by the RC techniques~\cite{YZhang2022}. One challenge is about the length of the guiding series. To retrieve the attractors encoded in the machine, the guiding series used for driving the reservoir in the listening stage is required to be sufficiently long~\cite{ZLu2020,Rohm2021,Roy2022,Gauthier2022}. As the system has already settled into the asymptotic attractors after such a long transient period, the adoption of machine-learning techniques is unnecessary. Another challenge is about overlapped attractors. To make the internal representations of the attractors in the reservoir distinctly different from each other, it is required that the coexisting attractors should be clearly separated in the phase space~\cite{ZLu2020}. Otherwise, if the attractors are overlapped, the one-to-one correspondence between the state of the target system and the state of the reservoir network will be lost, resulting in failed predictions. This requirement restricts seriously the application of RC, as in realistic systems the coexisting attractors are generally overlapped, e.g., the same set of neurons may participate in a variety of cognitive tasks in the neurocortex~\cite{MultiPerception,MultiMemory}. In what follows, we are going to show that both challenges are well overcome by the technique of balanced RC.   

\begin{figure}[tbp]
\begin{center}
\includegraphics[width=0.75\linewidth]{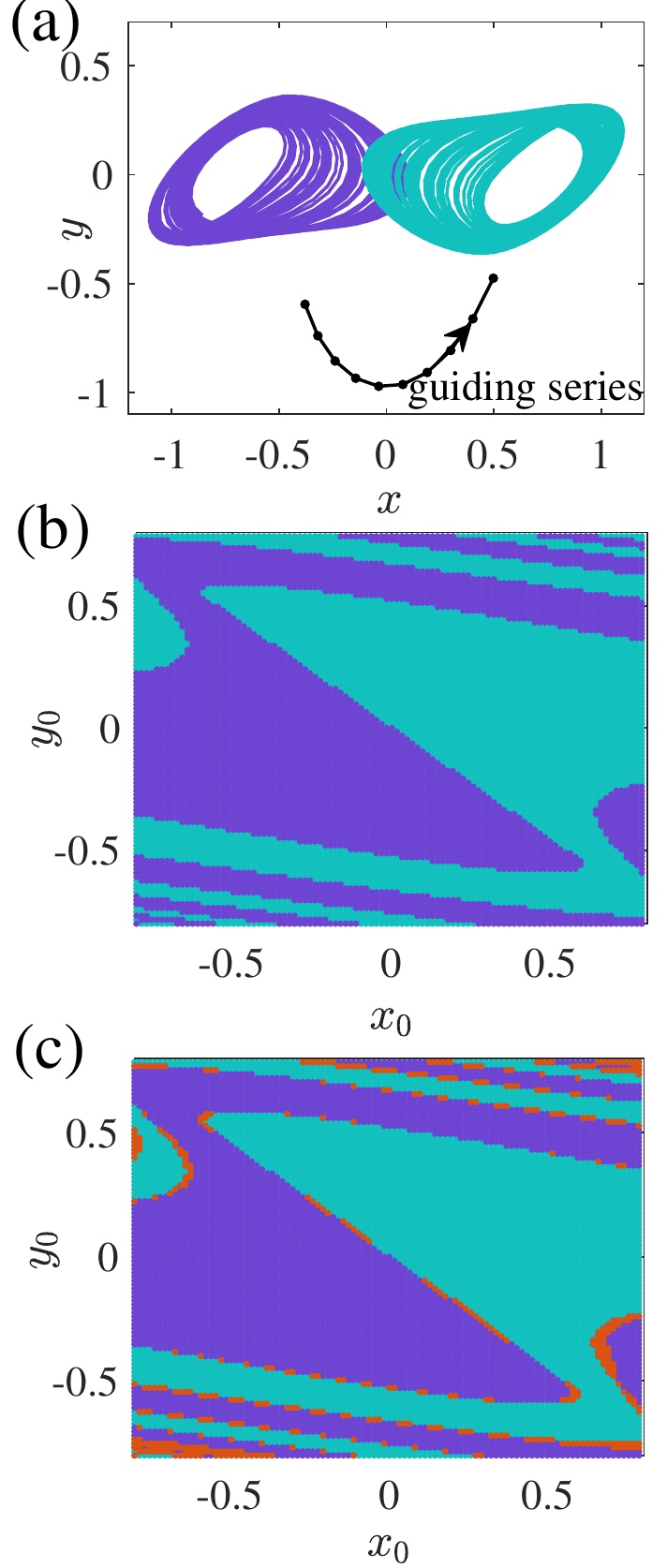}
\centering\caption{Inferring the attracting basins of coexisting chaotic attractors in chaotic Chua circuit. (a) The coexisting attractors. The attractor on the left and right sides are denoted as $\mathcal{A}_l$ and $\mathcal{A}_r$, respectively. Black dots denote the guiding series, which contains $l=10$ data points. (b) The attracting basins obtained by model simulations. (c) The attracting basins predicted by the machine. The prediction accuracy is about $96\%$. Red dots denote the false predictions.} 
\label{fig6}
\end{center}
\end{figure}

The first model we consider is the Chua circuit. The system dynamics is governed by the set of equations~\cite{ChuaModel}
\begin{equation}
\begin{cases}
\dot{x}=c_1[z-x-g(x)],\\
\dot{y}=c_2(x-y+z),\\
\dot{z}=-c_3y,
\end{cases}
\label{Chua}
\end{equation}
with $g(x)=m_1x+(m_0-m_1)(|x+1|-|x-1|)/2$ the piecewise-linear function. We set the system parameters as $(c_1,c_2,c_3,m_0,m_1)=(15.6,1,33,-8/7,-5/7)$, by which the system dynamics is chaotic and the largest Lyapunov exponent is about $\Lambda\approx 0.23$. In numerical simulations, Eq.~(\ref{Chua}) is solved by the 4th-order Runge-Kutta algorithm with the time step $\delta t=0.05$. Depending on the initial conditions, the system may evolve into different attractors. Shown in Fig.~\ref{fig6}(a) are the two attractors that the system might settle into, which are symmetric about the origin. Clearly, the two attractors are partially overlapped in the phase space. We denote the attractor on the left side as $\mathcal{A}_{l}$ (purple color), and the one on the right side as $\mathcal{A}_{r}$ (green color). Fixing the initial condition of variable $z$ as $0$, we plot in Fig.~\ref{fig6}(b) the attracting basins of $\mathcal{A}_{l}$ and $\mathcal{A}_{r}$ in the $(x,y)$-space based on the results of model simulations. Here the task to be accomplished by the technique of balanced RC is: given that the system evolution started from several initial conditions is measured, can we predict which attractor the system will settle into by noticing the initial evolution of the system started from a random initial condition? [An example of the initial evolution of the Chua circuit is plotted in Fig.~\ref{fig6}(a) (black dotted curve), which contains $l=10$ data points and sustains for a period of $T=0.5$ (about one-fiftieth of the system oscillation).]  

In implementing the prediction task, we first generate the training and testing data by recording the state evolution of the system started from $N=10$ different initial conditions, among which half of the initial conditions are developed to $\mathcal{A}_{l}$ and the other half are developed to $\mathcal{A}_{r}$. The variables $x$ and $z$ are normalized to be within the range $[-1,1]$, while the variable $y$ is kept unchanged. Each time series contains $\hat{L}=3000$ data points, and the training (testing) data contains $15000$ data points in total. For each time series in the training (testing) data, the first $l=20$ data points are used as the listening series for driving the reservoir out of the transient, and the remaining data points are used for machine training (validation). Still, we fix the size of the reservoir network as $n=500$, while optimizing the hyperparameters $(p,\lambda,\sigma,\alpha,\eta)$ in searching for the optimal machine. The ranges of the hyperparameters are identical to that of the power system, and the optimal machine is obtained after $400$ trials. According to the ratio of the prediction and synchronization errors, here we set the balancing parameter as $\beta=20$, for which the optimal hyperparameters are $(p,\lambda,\sigma,\alpha,\eta)=(0.7691,0.300,2.763,0.424,1.1\times 10^{-3})$.

In the predicting phase, the reservoir is first driven by the guiding series for $l=10$ data points, and then is operated in the closed-loop configuration for $L=10000$ steps. The system is regarded as having settled into $\mathcal{A}_{l}$ if the time-averaged variable $\bar{x}$ is negative for the last $1000$ predictions, and to $\mathcal{A}_{r}$ if $\bar{x}$ is positive. The attracting basins predicted by the machine are plotted in Fig.~\ref{fig6}(c). The prediction accuracy is about $96\%$. Comparing to the ground-truth results plotted in Fig.~\ref{fig6}(b), we see that, similar to the results of the power system [e.g., Fig.~\ref{fig3}(c)], the failed predictions are also distributed along the basin boundaries. [The impacts of intrinsic noise on the predictions have been also checked. Unlike the power system, the phenomenon of stochastic resonance is not observed here (not shown).]
%the regular distribution of false prediction is due to the piecewise linear feature of Chua circuit.

\begin{figure}[tbp]
\begin{center}
\includegraphics[width=0.75\linewidth]{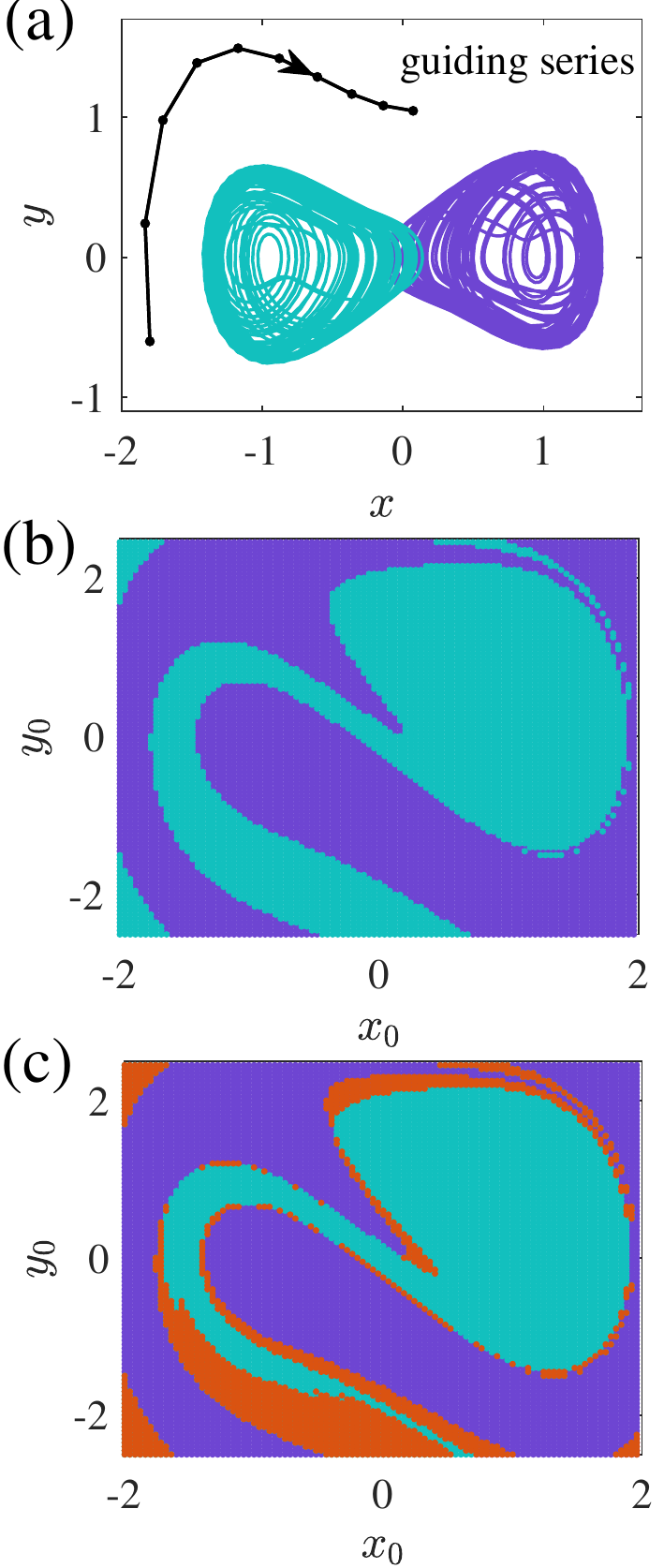}
\centering\caption{Inferring the attracting basins of coexisting attractors in the chaotic Duffing oscillator. (a) The coexisting attractors in the phase space. Black dots denote the guiding series, which contains $l=10$ data points. (b) The ground-truth results of the attracting basins obtained by model simulations. (c) The attracting basins predicted by the machine. The prediction accuracy is about $95\%$. False predictions are represented by red dots.} 
\label{fig7}
\end{center}
\end{figure}

The second model we adopt is the chaotic Duffing oscillator. The system dynamics reads
\begin{equation}
\begin{cases}
\dot{x}=y,\\
\dot{y}=-ky+x-x^3+A\sin(\Omega t),\\
\end{cases}
\label{Duffing}
\end{equation}
with $k$ the dissipation coefficient, $A$ the ampltude of the external driving, and $\Omega$ the driving frequency. Setting $(k,A,\Omega)=(0.5,0.38,1)$, the system presents the chaotic motion, with the largest Lyapunov exponent $\Lambda\approx 0.04$. Depending on the initial conditions, the system may evolve to different chaotic attractors, as depicted in Fig.~\ref{fig7}(a). We denote the attractors on the left (right) side as $\mathcal{A}_l$ ($\mathcal{A}_r$). It is seen that the two attractors are symmetric about the origin and are partially overlapped in the phase space. Based on the results of model simulations, we plot in Fig.~\ref{fig7}(b) the attracting basins of the two attractors, which shows that the basins are entangled and the boundaries separating the basins are irregular. Our mission here is to reconstruct the attracting basins utilizing the technique of balanced RC.

In this application, the training (testing) data is composed of $N=10$ time series, with each attractor contributing $m=5$ time series. The data are obtained by evolving Eq.~(\ref{Duffing}) numerically with the time step $\delta t=1\times 10^{-2}$. Each time series contains $\hat{L}=300$ data points, and the variables $x$ and $y$ are normalized to be within the range $[-1,1]$. In the training (validation) phase, the listening series contains $l=20$ data points. Again, we fix the size of the reservoir network as $n=500$, while optimizing the hyperparameters $(p,\lambda,\sigma,\alpha,\eta)$ in constructing the optimal machine. According to the prediction and synchronization errors, here we set the balancing parameter as $\beta=25$,  for which the optimal hyperparameters are $(p,\lambda,\sigma,\alpha,\eta)=(0.995,0.501,0.607,0.631,1.9\times 10^{-3})$ (obtained after $400$ trails). In the predicting phase, the guiding series contains $l=10$ data points [as depicted in Fig.~\ref{fig7}(a)], and the time series predicted by the machine contains $L=10000$ data points. The system is regarded as approaching $\mathcal{A}_l$ if $\bar{x}<0$ and approaching $\mathcal{A}_r$ if $\bar{x}>0$, with $\bar{x}$ being averaged over the last $100$ data points in the predicted time series. The predicted results are plotted in Fig.~\ref{fig7}(c), in which the false predictions are represented by red points. A comparison with the ground-truth results plotted in Fig.~\ref{fig7}(b) shows that the prediction accuracy is about $95\%$.  

\section{Discussions and conclusion}

A few remarks on the implementation and performance of balanced RC are in order:   

\begin{enumerate}[1)]
\item In preparing the training and testing data, we have collected time series from all the coexisting attractors (asymptotic states). This manner of data collection makes the machine capable of ``sensing" the global dynamics of the target system and thereby is necessary for reconstructing the attracting basins. This point has been confirmed by additional simulations, which show that if the data are collected from only some of the attractors, the performance of the machine will be significantly deteriorated (not shown). Furthermore, in composing the training and testing data, we have adopted from each attractor the same number of time series. This adoption is just for the sake of simplicity. It is possible that the machine performance can be further improved by adopting a different sampling strategy, e.g., choosing the number of time series for each attractor according to its basin volume. 

\item The performance of the machine is dependent on the length of the guiding series. While the prediction accuracy can be efficiently improved by increasing the length (time duration) of the guiding series, a shorter guiding series is more feasible and desirable from the perspective of real applications, due to the cost and difficulty in data acquisition. In particular, in infrastructure systems such as power grids, a shorter guiding series also means a more prompt response to dysfunctions and cascading events, which is critically important for system management. To demonstrate the efficacy of balanced RC, we have set the length of the guiding series in such a way that (1) it is much less than the period of the system oscillation and (2) no sign of the asymptotic state is shown during this episode. The adoption of a very short guiding series is the key difference between the technique of balanced RC and the multifunctional RCs proposed in the literature~\cite{ZLu2020,Rohm2021,Flynn2021}.

\item The machine performance can be further improved by optimizing the balancing parameter $\beta$. As depicted in Fig.~\ref{fig2} and expressed in Eq.~(\ref{error}), the role of $\beta$ is to balance the performances of state-evolution prediction and reservoir-network synchronization. As the synchronization performance is dependent on the length of the driving series (the longer the series, the smaller the synchronization error), the balancing parameter is also dependent on the driving series. In our studies, the value of $\beta$ is chosen according to the ratio between the prediction and synchronization errors, which is slightly different from the criteria used in predicting the attracting basins (e.g., based on the sign of the time-averaged variables). This makes it possible to improve further the performance of the machine by optimizing $\beta$, e.g., treating $\beta$ as an additional hyperparameter.

\item The inference of complicated attracting basins remains a challenge. For the models studied in our present work, the failed predictions are all observed at the boundaries of the attracting basins. This observation is understandable, as the system is sensitive to small perturbations nearby the basin boundaries. This raises the question of how to infer complicated basins, e.g., fractal or riddled basins, using the technique of balanced RC. One approach to solving the question could be increasing the length of the guiding series, which, however, will degrade the efficacy of the technique.     

\item The applicability of the new technique to other multistable systems, particularly the realistic systems, is yet to be checked. For demonstration purposes, the multiple dynamical models we have adopted are of low-dimension and simple attractors. It remains unknown whether the proposed technique can be applied to high-dimensional systems or systems with more complicated attractors, e.g., the attracting basins of the synchronization patterns in spatiotemporal systems of coupled oscillators. 

\end{enumerate}

To summarize, we have proposed a new machine-learning technique, namely balanced RC, and utilized it to infer the global stability of a power system. We have demonstrated that the new machine is able to not only predict whether the power system will return to its functional state in the presence of a large, random perturbation, but also reconstruct the attracting basin of the functional state with high precision. The new technique is featured by a very short guiding series in making the predictions, and the success of the new technique is attributed to the good balance between the performances of state-evolution prediction and reservoir-network synchronization. The impact of noise on predictions has been checked, and a stochastic-resonance-like phenomenon is observed. The new technique has been also applied successfully to infer the attracting basins of coexisting attractors in typical chaotic systems, including the Chua circuit and the Duffing oscillator. The results show that, though the coexisting attractors are overlapped in the phase space, the machine is still able to reconstruct the attracting basins with a high precision. The technique of balanced RC provides a powerful tool for the data-based stability analysis of power systems, and paves the way to the model-free prediction of multistable dynamical systems. 

\begin{acknowledgments}
This work was supported by the National Natural Science Foundation of China (NNSFC) under Grant Nos.~12275165 and 12105165. XGW was also supported by the Fundamental Research Funds for the Central Universities under Grant No.~GK202202003.
\end{acknowledgments}

\end{document}